# The performance evaluation of Multi-representation in the Deep Learning models for Relation Extraction Task

## La evaluación del desempeño de la multi-representación en los modelos de aprendizaje profundo para la tarea de extracción de relaciones.

Jefferson A. Peña Torres, Raul Ernesto Gutierrez,
Victor A. Bucheli, Fabio A. Gonzalez O.

**Abstract**

**Single implementing, concatenating, adding or replacing of the representations has yielded significant improvements on many NLP tasks. Mainly in Relation Extraction where static, contextualized and others representations that are capable of explaining word meanings through the linguistic features that these incorporates. In this work addresses the question of how is improved the relation extraction using different types of representations generated by pretrained language representation models. We benchmarked our approach using popular word representation models, replacing and concatenating static, contextualized and others representations of hand-extracted features. The experiments show that representation is a crucial element to choose when DL approach is applied. Word embeddings from Flair and BERT can be well interpreted by a deep learning model for RE task, and replacing static word embeddings with contextualized word representations could lead to significant improvements. While, the hand-created representations requires is time-consuming and not is ensure a improve in combination with others representations.**

## 1. INTRODUCTION

In recent years, models and techniques for automatically learning representations of text data have become an essential part of the natural language processing (NLP) works. For words to be processed by deep learning models, is need some form of representation. Commonly, a numerical or vector representation that can be calculated from a language representation model. Word Embeddings, One-hot Vectors and other vectors from representations models are a standard component of modern NLP architectures. The more simple vector representations is a one-hot encoded vector where categorical variables are represented as binary vectors, if are words, vector of size equal to vocabulary is setted with all zero values except the marked with a 1 which is a unique representation for each word. Although, other hand-written features as relative position of the words, POS tag can be encoded using these representation not incorporate additional information as similarity or context. In addition, the large size and sparse vectors no much used as main representation.

On the other hand, neural-based representation models such as Word2Vec [1], Glove[2] and FastText[3] that are among the best known models for generating static word embeddings, where a single vector, lacking of context have a same representation for each word in a vocabulary. These vectors follows the distributional principe, encode relationships between the semantic meanings of the words, the similarity behavior of the words. In recent works, namely deep neural language models where contextualized vectors can be generated with ElMo[4], Flair[5] are two relevant models. The words representation are sensitive to the context in which appear. In this study, we propose a extended model from a feed-forward neural network such as a CNN, replacing and concatenating static and contextualized representations for RE task.

At the same time, we show that by combine static and contextualized embeddings, we can make a comprehensive of word embedding models analysis and the features linguistic that these incorporate. Moreover, the experiments not only shed light on the properties of some embeddings for downstream task but can also serve as a approach for understanding existing biases in the pre-trained and the transfer knowledge through of the representations.

Our evaluation the impact of representations on deep learning model reveals posibles usages in another domains and languages as the medical where context, characters of the words, prefixes and suffixes are linguistic features that could be incorporated in a representation. The rest of this paper is organised as follows. Subsection 1.1 presents related work including research gaps and motivation for this paper. Section 2 presents representation models used in this paper together with the word embeddings configuration and benchmark corpora evaluated. In Section 3 we describe the experimental setup and in Section 4 experimental results and discussion. Finally, in Section 5 we conclude the paper.

## 2. RELATED WORK

In this work, we study the impact of multi-representations on a deep learning-based model for RE task, where commonly setting static variety. The key idea of models of representation is build vectors that capture quite a bit of the information of the words, its meaning, its associations with other words and so forth. The static representations models form part of language modeling, that follow the distributional hypothesis, the meaning of a word is systematically linked to the context in which it occurs [6], [7] . With a widespread acceptance we identify that Word2Vec [1], Glove[2] and FastText[3] as being are frequently chosen static embeddings not only because of the information about the word similarity that incorporate, but also there are pre-trained version on large dataset and publicly accessible in order to allow comparison across studies. Although, learn embeddings iteratively all implicity perform a word context-matrix from co-occurrence static[8] . This cause a unique representation for each word in the vocabulary, a problem due to the nature of the words and the language, the polysemous is not capture by a static vector.

Similarly, in considering different notions of context, recent neural architectures has allowed contextualized representations models which has increased the ability to include more information of the word in its vector representation. Models as BERT[9], ELMo [4], FLair[5], GPT[10], and so forth, achieve multimodal context, which capture the meaning of words in general and also its understanding in the context where occurs[11] . Taking into account the limitation of the static word representations, the success of contextualized suggests that these representations capture highly transferable and task-agnostic properties of language[12], [13] .

Although, word representations are common in deep learning there are several hand-created features that can be represented. These representations are obtained directly from the data with which one works without having to train a model previously. we call these as simple representations because are naive, one-hot representation does not provide information about word relationships, and presents dimensionality problems, but could be used for incorporate informatión of position of word in sentence, POS tag of the word and so forth. For our analysis, we setting a baseline model on the premise that a ready trained model could reach better results if a right selection of features incorporates in the representation are learned.

## 3. MODELS AND METHODS

We evaluate the performance of a Relation Extraction (RE) model CNN-based architecture using the measures of precision, recall and f1-score, the input sentences to feed the model are encoded separately as a representations made up of word embeddings and other vector representations. F1 score was considered as an evaluation metric rather than being a measure of accuracy and as the measures were highly variables, with box plots we visualize the distribution of values and the outliers of f1-scores reached by the model with each representation. In this section, we describe dataset details, model implementation, performance measures and word representations models used.

### 3.1. Datasets Details

To analyze representations, we use Word2vec, Glove and Fastext word representation trained on a common dataset, Wikipedia and a pre-trained version of ELMo, Flair, Pooled and Character as contextualized models. Our input data to RE model come from the SemEval 2010, task 8 by [14]hen, a common dataset for RE task that in which the sentences and nominals has been labeled. For example,

PART-WHOLE relation relation between *car* and *engine* in *"the car has an engine"*. Or a CONTENT-CONTAINER relation in the sentence *"The bowl contained apples, pears, and oranges."*

If the RE model use the information from representation and improve its performance, we could infer that there was a incorporate feature that the representation model take into account for build word representations. Using these datasets, we train the RE model and representations. We consider train, dev and test sub division for evaluation and measures extraction.

### 3.2. Model Implementation Details

The CNN architecture learn informative patterns present in representations, using three convolutional filters of size 3, 4, 5 applied on a embedding matrix. To train a CNN-based model a word representations and other vector representations sequence are given as input, a serie of vectors that make up for each sentence an embedding matrix. While training the model extract automatically features and the output is a vector of length equal to number of relation types in dataset.

Output vector has the probability values that indicate which relationship has in the input. In this work, nine relation types are considered into data set selected. The model consist of multiple layers, a convolutional layer, a max pooling, a feedforward and a fully connected layer. The model implement as baseline for our experimentation is similar to Nguyen, specifically, we make the assumption that a ready build model could be improved just with right representation.

### 3.3. Performance Measures

We measure the performance using three common metrics: *precision* (P), *recall* (R) and *f1-score* (F1) for train/validation dataset on our RE model. Although, the f1-score was used as main metric, because is the harmonic average of the corresponding mean precision and recall values and moreover, we analyze the F1 score according to features incorporated in the representations considered.

$$P = TP/(TP+FP) ; R = TP/(TP+FN)$$
$$F1 = 2(P*R)/P+R$$

We trained our networks using Adam optimization with a learning rate of 0.001 and a batch size of 50. We trained the contextualized representations with 70 epochs and others with 120 epochs.

### 4. EXPERIMENTS AND RESULTS

In this section, we empirically evaluate the performance of a RE model, where representation of words and other features from text in vector form for are the first pillar our experimentation: they provide the representational basis to turn texts into matrix, and play a central role in the pattern learning automatic. We also provide insights on the impact of the some common representation models and we quantified our model performance in terms of precision, recall and F1 score (also called F-measure).

There are a number of stochastic elements in the training session for a deep learning model, which makes exact replication of the results difficult, particularly performance results. Instead, in this work each representation include in model is trained and tested 10 times with different random seeds resulting in a span of precision, recall and F1 scores. Into our tables the minimum value reached by each representation and on more informative boxplots mean, maximum, minimum, outliers and so forth values of f1-score.

| Static Word embeddings | | | |
|---|---|---|---|
| **Model + representation** | **P** | **R** | **F1** |
| Word2vec 50d | 74.34 | 79.67 | 79.16 |
| Word2vec 100d | 76.06 | 78.48 | 79.58 |
| Word2vec 300d (*) | 79.49 | 81.21 | <u>80.28</u> |
| Glove 50d | 73.25 | 79.25 | 76.06 |
| Glove 100d | 76.16 | 80.32 | 78.15 |
| Glove 300d | 77.67 | 80.64 | 79.10 |
| Fastext 300d | 73.15 | 74.34 | 73.67 |
| **Contextualized Word embeddings** | | | |
| **Model + representation** | **P** | **R** | **F1** |
| Flair | 80.18 | 83.25 | 82.64 |
| ELMo | 79.74 | 83.16 | 82.41 |
| CharEm | 79.45 | 80.15 | 80.58 |
| Pooled | 76.40 | 78.82 | 77.50 |
| BPR | 70.23 | 76.41 | 73.18 |

**Table No 1.** F-measure scores obtained in 10 times execution of the model on SemEval dataset. For convenience, the F-measure benchmark is reported in underline in the last row. Column P and R gives the number of precision and recall performed while holding the model and representation in first column.

In Table No 1. we summarize, static, contextualized, and combination with other vectors representations. With a top F1 score of 81.64 % reached for the extend model, our results show clearly outperforms previous DL model, with a 79.58 % F1. The top performances of baseline is reached with static representation with consistently lie below the average performance level attained by our extended model.

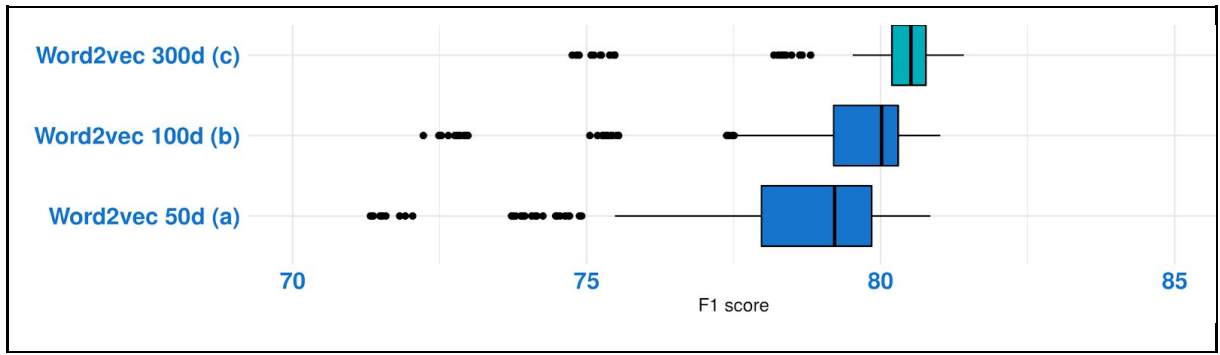

**Figure No 1.** Boxplots of F-measure scores by representation dimension values. Boxplots related to different vector dimension from Word2vec the lower share the same color. This plot illustrates the variation of f1-score by each tested vector dimension on RE model

A part of information provided in Table 1 is conveyed visually in Figure 1, which shows boxplots of F-measure scores obtained different vector dimensionality. Dimension of the dense vector indicates the vector space in which words will be embedded and for our experiments we used 50, 100 and 300, which are commonly selected in works as, the evidence with Word2vec vectors show that a major dimensionality on the model could also trigger achieved a better result, we also setting vector with same dimension from Glove and similar evidence was obtained.

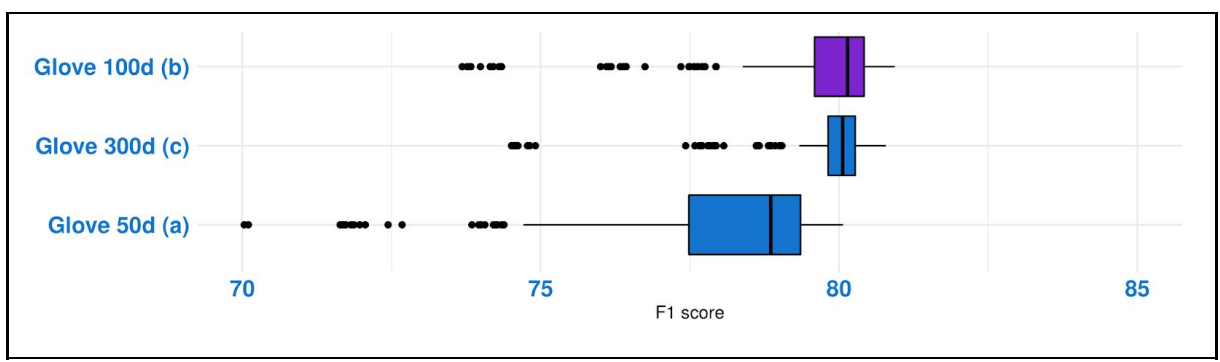

**Figure No 2.** Boxplots of F-measure scores by representation dimension values. Boxplots related to different vector dimension from Glove the lower share the same color. This plot illustrates the variation of f1-score by each tested vector dimension on RE model

In Figure 2, word embeddings from Glove show improvements with a high dimensionality, but no with the highest dimensionality considered. This f1-score gap can be explained by reference to the fact that in word2vec neural networks are widely used for vectors buildings and with a high-dimensionality, this include more information of the words, for other way, glove which not incorporate neural networks but if a statistic principle and global information of the word in a high dimensionality not incorporate more information about of the words. Moreover, distributional principle and features such as context of individual words extracted without human intervention make accurate information of the word, this is important feature for relation extraction task. Although in a practical sense, both model on our model reach almost similar performance is reached.

An important choice for the model performance is the dimensionality, the evidence show that a major dimensionality may cause the scores to vary less. All our models are benchmarked with dimension vectors 300 as the baseline model is highlighted in Table 1. The RE model also too have other representations, the word are key in relation extraction and others works has implemented features as subtrees[15], relative position of words respect to nominals[16], [17] among others features that can be manually extracted, we consider POS tag and characters as features for feasible representation because the baseline model already incorporates position of the nominals.

| Features as representations | | | |
|---|---|---|---|
| Model + feature | P | R | F1 |
| Position Embeddings | 79.49 | 81.21 | **80.28** |
| POS Tag Embeddings | 75.32 | 79.90 | 77.41 |
| Characters Embeddings | 74.09 | 79.20 | 76.53 |

**Table No 2.** F-measure scores obtained in 10 times execution of the model on SemEval dataset. For convenience, the F-measure benchmark is reported in underline in the last row. Column P and R gives the number of precision and recall performed while holding the model and additional feature in first column.

In Table 2 summarize the results with features as representations, one factor for representation usage is that deep learning models automatically learn relevant patterns. A good representation of the raw data carries out an automatic feature engineering and additionally the vectors from these representation could be enriched with additional features as position embeddings which indicate the distance among words in sentence and each nominal. In the feature engineering for RE task the position the words close to the entities are informatives to determine the relation.

We also replaced and combined features and embeddings both as vectors, and the evidence suggest to try several combination of features to select best combination to give a high performance. It is simply not appropriate, because need hand-created features and external resources with previously training. The position perform well because is extracted from sentence when nominals has been identified, the POS tag requires external tool as CoreNLP[17], using a sparse vector representation. Also, the Characters do not have a standard dense representation although, there is approaches as in [18], but, on the model does not a high performance that the position feature.

As important feature in word embeddings is the context, deep contextualized word representations incorporates complex characteristics of the word, mainly syntax and semantics features which are automatically learned by several neural networks layers.

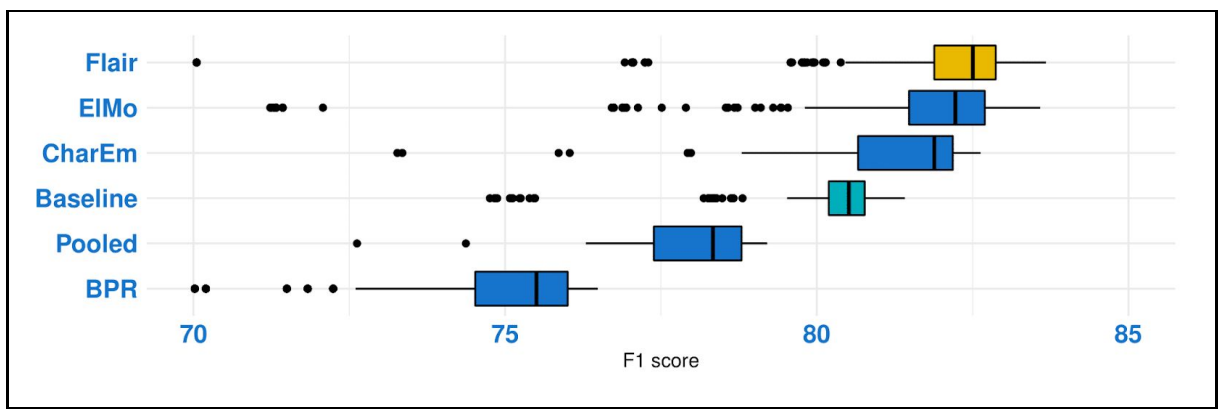

**Figure No 3.** Boxplots of F-measure scores by representation dimension values. Boxplots related to different vector dimension from Contextualized models the best in yellow, baseline in green and others share the same color. This plot illustrates the variation of f1-score by each contextualized representation on our model.

We summarize the contextualized representations on our model, in Table 1 and Figure 3. As a effective replacement for static word representations, contextualized vectors from FLair model reach a better performance. The combination with position embeddings achieve the best results, although, each representation model incorporated stand-alone features, some improves the model as in Flair where combination of statical representation from FasText (3), character-level features and contextualized representation from(4,18) on Bi LSTM-CRF architecture generates more informative vectors than single ELMO architecture. The character information is a important feature for vector contextualization, in FLair, ELMO and CharEm is included, these reach results greater than benchmark model. Others contextualized models that include other features as in Pooled and BytePair not so well behaved in our models.

The combination, concatenation and replace are operations that could be used for improvements of any model. We concatenate contextualized, static and feature vectors for a more informative input.

| Multi-Representation | | | |
|---|---|---|---|
| Model + representation | P | R | F1 |
| Baseline | 79.49 | 81.21 | **80.28** |
| Baseline + Flair | 79.17 | 82.32 | 80.43 |
| Baseline + Elmo | 79.82 | 83.59 | 81.66 |
| Baseline + CharEm | 78.83 | 82.13 | 80.71 |
| Baseline + Pooled | 76.99 | 79.78 | 78.30 |
| Baseline+BPR | 76.40 | 81.45 | 78.83 |

**Table No 3.** F-measure scores obtained in 10 times execution of the model on SemEval dataset. For convenience, the F-measure benchmark is reported in underline in the last row. Column P and R gives the number of precision and recall performed while holding the model and additional feature in first column.

In Table 3 the performance of our models, note that context and character are features that are take into account into these models for generate contextualized vectors. This multi-representation outperform the single static representation and even the combination static and feature representation, with ELMO, Word2vec and position embeddings out model reach the best result. In same way, Figure 4 show the performance reached by our models, the concatenation of elmo word embeddings, wor2vec word embedding and position embeddings outperform the patterns that the model learn. In general multi-representation improve the results on our model, all contextualized vector representation in concatenation in comparison with single replacing of static by contextualized vectors. Although, the outliers in the plots indicates increasing in the variability of f1-scores.

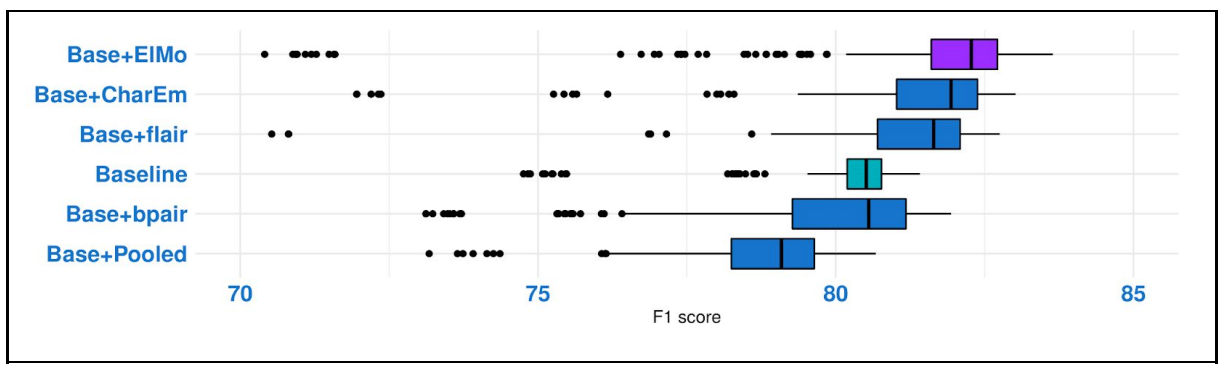

**Figure No 4.** Boxplots of F-measure scores by representation dimension values. Boxplots related to different vector dimension from Contextualized models the best in purple, baseline in green and others share the same color. This plot illustrates the variation of f1-score by multi-representation on our model.

## 5. CONCLUSION

In order to determine which word vector model to use, each model were evaluated using a set of experiments where static, contextualized and other representations are considered. We setting a already deep learning model for RE and with a common standard dataset proof Word2vec, Glove, Fasttext as static representations models, ELMO, Flair, Pooled Embeddings, BytePair and Character Embeddings as contextualized models.

Contextualized word vectors produces results considerably better than static word vector approaches, with baseline model performance equal to 79.82%, with FLair a increase of f1 score to 81.64% with FLair is reached, replacing static with contextualized word vectors whereas, ELMo increase to 81.66 % if there is a combination of these of contextualized and static.

Empirical evidence shows that our proposal models outperform when right representation is choice for the know task. Our approach exhibits desirable properties for representation selection, (i) The dimensionality. Useful linguistic features for any NLP task are learned by the DL model automatically, a major vector can lead a minor variation of performance. Although, specific results may vary given the stochastic nature of the deep learning approach. (ii) Additional features. With only incorporate knowledge in representation a DL model identify, classify or extract according to specific task. With features as vector representation the model used obtain more information of specific task, an important improve for the relevant patterns learning from representations. (iii) Multi-representation. Several operations as concatenation, replace and so forth could be reach better results. Deep contextualized models use widely neural networks, combination other representations, features and architectures as LSTM to build vectors that dynamically represent word in its context.

At the moment, we are running experiments in which a CNN-based model pre-trained with static word representation is trained to classify using multi-representations. The Transfer Learning from pretrained allows researchers discover explicit and implicit features incorporate in the model. Although, no is a easy task, set connections between representation and performance, could be considered as open problem of research where is likely to continue to produce models for word representation and models that take advantage of these and reach better performance.

# 6. REFERENCES

[1] T. Mikolov, I. Sutskever, K. Chen, G. S. Corrado, y J. Dean, «Distributed representations of words and phrases and their compositionality», en *Advances in neural information processing systems*, 2013, pp. 3111–3119.

[2] J. Pennington, R. Socher, y C. Manning, «Glove: Global vectors for word representation», en *Proceedings of the 2014 conference on empirical methods in natural language processing (EMNLP)*, 2014, pp. 1532–1543.

[3] P. Bojanowski, E. Grave, A. Joulin, y T. Mikolov, «Enriching word vectors with subword information», *Trans. Assoc. Comput. Linguist.*, vol. 5, pp. 135–146, 2017.

[4] M. E. Peters *et al.*, «Deep contextualized word representations», *ArXiv Prepr. ArXiv180205365*, 2018.

[5] A. Akbik, D. Blythe, y R. Vollgraf, «Contextual String Embeddings for Sequence Labeling», en *COLING 2018, 27th International Conference on Computational Linguistics*, 2018, pp. 1638–1649.

[6] J. R. Firth, «A synopsis of linguistic theory 1930-55.», *Stud. Linguist. Anal. Spec. Vol. Philol. Soc.*, vol. 1952-59, pp. 1-32, 1957.

[7] S. Deerwester, S. T. Dumais, G. W. Furnas, T. K. Landauer, y R. Harshman, «Indexing by latent semantic analysis», *J. Am. Soc. Inf. Sci.*, vol. 41, n.º 6, pp. 391–407, 1990.

[8] O. Levy y Y. Goldberg, «Dependency-based word embeddings», en *Proceedings of the 52nd Annual Meeting of the Association for Computational Linguistics (Volume 2: Short Papers)*, 2014, pp. 302–308.

[9] J. Devlin, M.-W. Chang, K. Lee, y K. Toutanova, «Bert: Pre-training of deep bidirectional transformers for language understanding», *ArXiv Prepr. ArXiv181004805*, 2018.

[10] A. Radford, K. Narasimhan, T. Salimans, y I. Sutskever, «Improving language understanding by generative pre-training».

[11] E. Bruni, N.-K. Tran, y M. Baroni, «Multimodal distributional semantics», *J. Artif. Intell. Res.*, vol. 49, pp. 1–47, 2014.

[12] I. Tenney *et al.*, «What do you learn from context? probing for sentence structure in contextualized word representations», *ArXiv Prepr. ArXiv190506316*, 2019.

[13] N. F. Liu, M. Gardner, Y. Belinkov, M. Peters, y N. A. Smith, «Linguistic knowledge and transferability of contextual representations», *ArXiv Prepr. ArXiv190308855*, 2019.

[14] I. Hendrickx *et al.*, «Semeval-2010 task 8: Multi-way classification of semantic relations between pairs of nominals», en *Proceedings of the Workshop on Semantic Evaluations: Recent Achievements and Future Directions*, 2009, pp. 94–99.

[15] S. Lim y J. Kang, «Chemical–gene relation extraction using recursive neural network», *Database*, vol. 2018, 2018.

[16] D. Zeng, K. Liu, S. Lai, G. Zhou, J. Zhao, y others, «Relation classification via convolutional deep neural network», 2014.

[17] C. Manning, M. Surdeanu, J. Bauer, J. Finkel, S. Bethard, y D. McClosky, «The Stanford CoreNLP natural language processing toolkit», en *Proceedings of 52nd annual meeting of the association for computational linguistics: system demonstrations*, 2014, pp. 55–60.

[18] C. N. dos Santos y V. Guimaraes, «Boosting named entity recognition with neural character embeddings», *ArXiv Prepr. ArXiv150505008*, 2015.